\documentclass[11pt,a4paper]{article}
\usepackage[hyperref]{acl2020}
\usepackage{times}
\usepackage{latexsym}

\usepackage{graphics}
\usepackage{graphicx}
\usepackage{csquotes}
\usepackage{makecell}
\usepackage{microtype}

\aclfinalcopy 

\title{FairPy: A Toolkit for Evaluation of Prediction Biases and their Mitigation in Large Language Models}

\author{ Hrishikesh Viswanath\\
  Dept. of CS \\
  Purdue University \\
   \\
  \texttt{hviswan@purdue.edu} \\\And
  Tianyi Zhang \\
  Dept. of CS \\
  Purdue University \\
   \\
  \texttt{tianyi@purdue.edu} \\}

\date{12}

\begin{document}
\maketitle
\begin{abstract}
Recent studies have demonstrated that large pretrained language models (LLMs) such as BERT and GPT-2 exhibit biases in token prediction, often inherited from the data distributions present in their training corpora. In response, a number of mathematical frameworks have been proposed to quantify, identify, and mitigate these the likelihood of biased token predictions. In this paper, we present a comprehensive survey of such techniques tailored towards widely used LLMs such as BERT, GPT-2, etc. We additionally introduce Fairpy, a modular and extensible toolkit that provides plug-and-play interfaces for integrating these mathematical tools, enabling users to evaluate both pretrained and custom language models. Fairpy supports the implementation of existing debiasing algorithms. The toolkit is open-source and publicly available at: \href{https://github.com/HrishikeshVish/Fairpy}{https://github.com/HrishikeshVish/Fairpy}.
\end{abstract}

\section{Introduction}

Large pretrained language models have emerged as powerful tools for capturing the complexities of natural language and interpreting textual data. They have demonstrated strong capabilities in generating syntactically and semantically coherent text that is contextually appropriate. These models are typically trained on massive corpora collected from internet sources. However, these models reflect the statistical distributions of the datasets, often leading to preferential token generations, following the semantics of the datasets \cite{bolukbasi2016man, kirk2021bias, papakyriakopoulos2020bias}, which affects the reliability of NLP systems and limits their adoption in real-world applications.

While previous studies have established the presence of prediction biases in pretrained models, the metrics proposed for bias detection are often theoretical in nature or tailored to narrow experimental settings \cite{jin2020transferability}. These methods are not always accessible, nor are they readily integrable into common development pipelines—posing a significant barrier for NLP practitioners. To address this gap, we introduce Fairpy, a Python-based open-source toolkit that consolidates a wide range of bias detection and mitigation techniques. Fairpy offers modular interfaces that support seamless integration with widely-used open-source language models as well as user-defined models. Additionally, it includes curated datasets and templated corpora commonly used in bias evaluation, enabling comprehensive and reproducible analysis.

The bias detection methods surveyed in this paper can broadly be categorized into two groups. The first category comprises metrics that assess biases in the model’s internal representations or encodings \cite{bolukbasi2016man, papakyriakopoulos2020bias}, where the learned embeddings of the model's vocabulary exhibit statistical regularities that correspond to the biases in the training data. The second category includes behavioral metrics, which infer bias by evaluating the model’s output in response to controlled input perturbations \cite{kirk2021bias, zhao2018gender, guo2021detecting, magee2021intersectional, li2020unqovering}.

Bias mitigation methods are similarly categorized into two main classes. The first class involves retraining the model using a balanced or curated dataset \cite{barikeri2021redditbias, bartl2020unmasking}, or adapting the model through reinforcement learning or transfer learning techniques \cite{park2018reducing, liu2021mitigating, jin2020transferability}. The second class focuses on post hoc interventions, such as modifying the learned token embeddings or internal representations to reduce biased associations \cite{qian2019reducing, faal2022reward}.

While many bias detection metrics and mitigation techniques have shown effectiveness on specific language models, their applicability often does not generalize across LLM architectures. This limitation primarily arises from the dependence on template structures used for evaluation, as well as the differences in internal representations across model types.

Causal language models are inherently generative, employing unidirectional attention mechanisms to predict the next token in a sequence based solely on preceding context. As a result, they are typically evaluated using complete, naturalistic sentence templates. In contrast, masked language models utilize bidirectional attention, leveraging both preceding and succeeding context to infer missing tokens. Consequently, they are commonly assessed using cloze-style templates with masked tokens in intermediate positions. Such masked templates are incompatible with generative models, limiting the portability and generalizability of bias metrics designed specifically for masked architectures.

To address these limitations, we propose a framework for constructing more adaptable template structures in future work, enabling unified bias evaluation across different language model types.

In this paper, we conduct a comprehensive discussion of existing bias detection and mitigation techniques. Furthermore, we modularize and decouple these techniques to enhance their generality, enabling broader applicability across a diverse range of large language models.

Our main contributions are -  
\begin{itemize}
    \item Decoupling bias detection metrics from language models and specific contexts and making them generalized.
    \item Presenting python interfaces to allow users to plug-and-play their own models to test these biases and a toolkit that compiles most of the available bias detection metrics and bias mitigation techniques
    \item Discussion of the techniques and their applicability. 
\end{itemize}

\section{Related Work}

\textbf{Measuring Bias in Token Predictions}

\cite{bolukbasi2016man} presented seminal work in quantifying bias in word embeddings by utilizing cosine similarity to measure directional relationships between tokens. They define bias as the non-equidistant positioning of embeddings relative to a neutral embedding for semantically opposite tokens, where the distances between semantically similar tokens are disproportionately skewed.

In their work, \cite{papakyriakopoulos2020bias} identify three types of biases in word embeddings: pre-existing biases, which arise from the input corpora; technical biases, stemming from the architecture and mathematical constraints of the model; and emergent biases, which manifest when the model makes predictions. The embeddings used in this study were static, generated with GloVe. To assess bias, they applied the Mann-Whitney U test and the Kruskal-Wallis H test on text generated for different semantic groups.

\cite{zhao2018gender} introduced the WinoBias dataset, demonstrating that pro-stereotypical entities were predicted with higher accuracy in contextualized scenarios.

\cite{guo2021detecting} present a comprehensive evaluation of emergent biases in language models such as BERT, GPT-2, GPT, and ELMo. They introduce the CEAT (Contextualized Embedding Association Test) to assess biases in contextualized embeddings, as opposed to static embeddings.

\cite{li2020unqovering} design ways to find confounding factors affecting bias of large transformer based language models. They test whether the predictions depend on the position of the subject or if the predictions change if an attribute is negated. They also provide a framework to determine biases in underspecified questions and seemingly uncorrelated sentences. The metrics were tested on BERT and its variants. 

\textbf{Biases in Machine Translation}
\cite{sun2020automatic} propose a method for mutating contextual words in input templates to observe how changes affect the un-mutated portion of the output. The resulting translations are evaluated for grammatical fluency, and predictive probability along with cosine similarity scores are used to assess bias in the output.

\cite{prates2020assessing} analyze biases present in Google Translate, providing a critical assessment of its translation outputs across various languages.

\cite{stanovsky2019evaluating} conduct the first large-scale multilingual bias detection study using the WinoBias dataset to identify biases in machine translation systems. Their approach involves translating words into different languages and aligning the translated terms with their English counterparts to detect bias across languages.

\textbf{Bias Reduction in Language Models}
\cite{park2018reducing} Evaluated Bias mitigation strategies on older NLP models such as CNN and GRU. They performed transfer learning on debiased corpora and debiased word embeddings by augmenting the existing corpora.  

\cite{jin2020transferability} Explore the concept of upstream mitigation as a means to debias BERT. They discuss whether it is possible to fine-tune against multiple biases in a single downstream tasks and provide a transfer learning framework to finetune and mitigate a single bias factor in upstream stage. 
\cite{qian2019reducing} Provide methods to remove bias in word-level settings, specifically with Glove Embeddings. They propose a loss function to equalize the probability of predicting words of a particular group. 

\cite{bartl2020unmasking} Experiment with reducing BERT's bias by retraining the model with the GAP corpus. They use real world statistics and cross linguistic approaches to build a bias evaluation template dataset. They also show that masked models exhibit bias in contextualized settings. 

\cite{faal2022reward} Analyze two techniques for detoxifying lanuage models - Data driven and decoding based techniques. Data driven methods involve augmenting data and retraining the model. The decoding based methods involve having a model that decodes the embeddings in an unbiased way as a form of post-processing. 

\textbf{Toolkits for measuring biases in Language Models}
\cite{ribeiro2020beyond}, in their work propose a task agnostic methodology called CHECKLIST to test NLP models for general linguistic capabilities. The tool performs three types of tests - The minimum Functionality test, which comprises of a set of small neutral sentences with simple adjectives. The invariance test performs label preserving perturbations to the template to check if the behavior of the NLP model differs. Lastly, the model uses Directional Expectation Test which adds a sentiment to the template and checks if the model doesn't predict the opposite sentiment. The metrics provided in this work form the basis for bias detection through counterfactual data augmentation but these metrics were not applied to pretrained language models.

\cite{geva2022lm} have built an interactive tool called LM-Debugger that illustrates the internal workings of Language Models. They note that most metrics follow probing or perturbation and underscored the need to understand the internal working of these models. At each step a set of tokens are provided and the users are allowed to choose a token. This choice determines the prediction at the following layers. The trace of the model helps determine how the biases are formed. 
\cite{nozza2022pipelines} present a bias detection Engine for measuring biases and integrating them in development pipeline. They propose a badge system to perform continuous evaluation of biases in CI/CD setting during the development of software that uses language models. 

\section{System Overview}
In this section, we discuss the architecture Fairpy, a model that provides flexible interfaces for applying bias detection and mitigation techniques on existing and custom language models. The model has two packages, the bias detection package and the bias mitigation package. The metrics defined in these packages currently support the following language models provided in HuggingFace Library - CTRL, GPT-2, GPT, TransfoXL, BeRT, DistilBeRT, RoBeRTA, XLM, XLNet, AlBeRT. The model also has interfaces to plug in user-defined or custom language models, along with their tokenizers. 
The Bias Detection package currently provides the following metrics - Hellinger Distance \cite{beran1977minimum}, WEAT Score \cite{caliskan2017semantics}, StereoSet Score \cite{nadeem2020stereoset}, Honest Score \cite{nozza2021honest}, Log Probability \cite{nangia-etal-2020-crows} and F1 Score. The Bias Mitigation package currently supports Dropout Debias, NullSpace Projection, Sentence Debias, DiffPruning, Self Debias and Counter Factual Data Augmentation. 
 To our knowledge, this is the first toolkit to provide such an interface. The package is available to download at \href{https://github.com/HrishikeshVish/Fairpy}{https://github.com/HrishikeshVish/Fairpy}
\section{Bias Detection Metrics}
This section describes the Bias Mitigation techniques that have been currently included in the toolkit.
\begin{figure*}[h]
    \centering
    \includegraphics[width=1\textwidth]{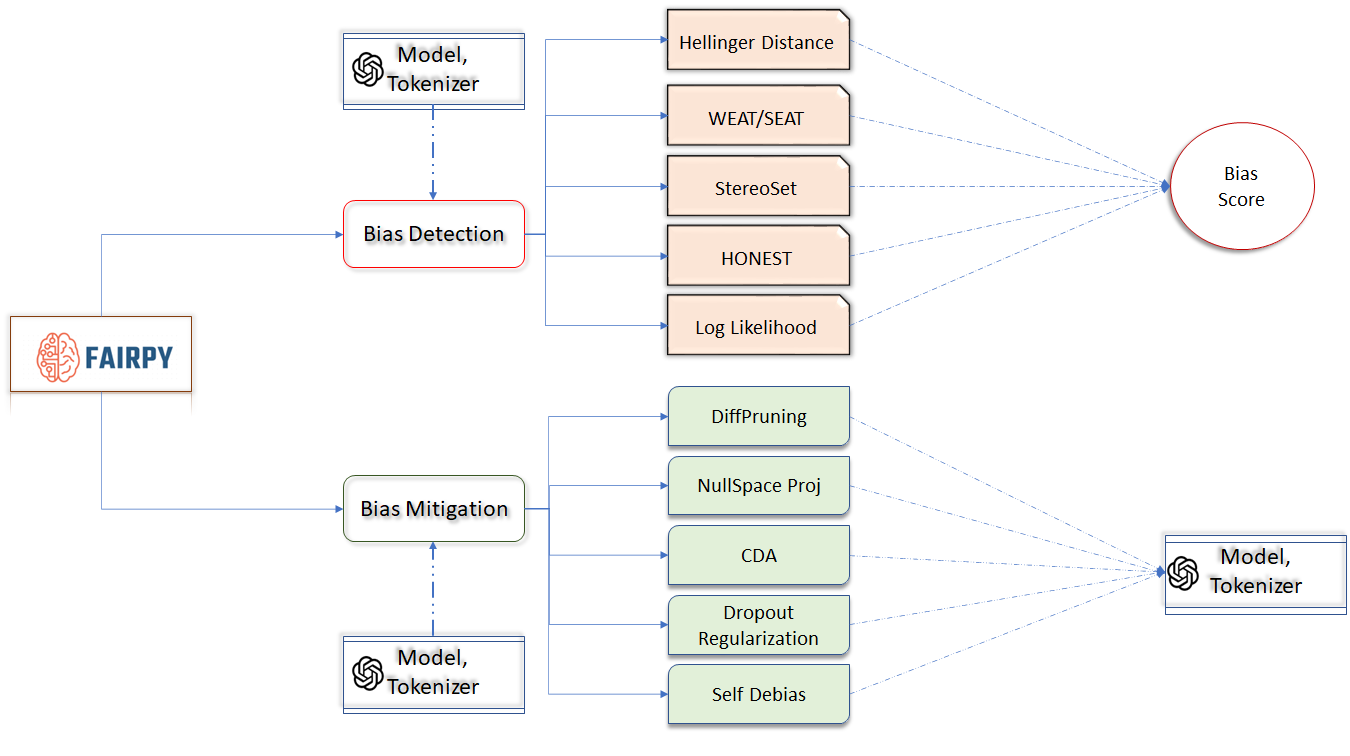}
    \caption{The flow diagram highlights the overall architecture of the model, with the nodes labelled in red denoting the bias detection techniques and the nodes in green highlighting the bias mitigation techniques}
    \label{fig:my_label}
\end{figure*}
\subsection{Hellinger Distance}
As presented in \citet{zhu2012information}, Hellinger distance metric is used to measure the difference between two probability distributions. Let us define two probability distributions P and Q such that P = $\{ p_i\}_{ i\epsilon[n] }$ and Q = $\{ q_i\}_{i\epsilon[n]}$. The Hellinger Distance between them is defined as follows \begin{equation}
    h(P, Q) = \frac{1}{\sqrt{2}} . || \sqrt{P} - \sqrt{Q} ||_2
\end{equation}
The Hellinger distance is computed between the embeddings generated by the language model for two different classes. The input corpus is augmented such that the nouns and pronouns are replaced with members of a particular class. The augmented corpus is fed to the language model and the output is extracted. This forms the context for the semantic class. The dot product of this context with the embedding forms the logits. Softmax is applied on this logits to get the probability distribution. The Hellinger distance between these probability distributions provides a measure of how biased the model is towards a particular class. 
\subsection{WEAT Score}
Word Embedding Association Test is a method proposed by \cite{caliskan2017semantics} and is effective in measuring the cosine similarity between contextualized embeddings generated by language models. WEAT score is computed by partitioning the sets of words into two classes. Let these be represented as M and F, the attribute classes. The target words are partitioned into two sets X and Y. The WEAT score for X and Y, with respect to M and F is given by the following equation 
\begin{equation}
    s(X,Y,M,F) = \sum_{x\epsilon X} a(x, M, F) - \sum_{y\epsilon Y} a(y, M, F)
\end{equation}
\begin{equation}
    a(w, M, F) = avg_m cos(\vec{w}, \vec{m}) - avg_f cos(\vec{w}, \vec{f})
\end{equation}
where $m\epsilon M$ and $f\epsilon F$ and a refers to the association between the word and the classes.
The effect size is given by 
\begin{equation}
    \frac{\mu_x a(x, M, F) - \mu_y a(y, M, F)}{\sigma_w a(w, M, F)}
\end{equation}

This toolkit uses the extension of the WEAT metric called SEAT or Sentence Encoder Association Test, as defined in \cite{may2019measuring} and implemented in \cite{meade2021empirical}. The embeddings for the words of these classes are compared against a comprehensive set of common nouns and adjectives with positive and negative implications. 

In the ideal case, the embedding representation of each word in the vocabulary is expected to be equidistant from the two attribute classes. Any deviation suggests bias in one direction. Greater the deviation, greater the bias. 

This metric is suitable for classes which are binary. However, effort has been made to extend them to multiclass scenario. One possible way that has been explored in this toolkit is to break down multiclass labels into binary. For example, while comparing nationalities, the classes would be Romanian vs Non-Romanian. 
\subsection{StereoSet Score}
StereoSet, proposed by \cite{nadeem2020stereoset} is a comprehensive dataset of English sentences and a set of tools for measuring biases. The model measures the likelhood of predicting words in sentences with masks in two settings - intersentence and intrasentence. 

The model comprises two scores - Language Model score and Stereoset score. The Language Model score gives higher score to meaningful associations. This score represents the percentage of associations that are meaningful. 

The stereoset score determines the percentage of examples in which the model prefers stereotypical associations over anti-stereotypical associations. The ideal score is 50, where half the examples are stereotyped and the other half aren't.

\subsection{Honest Score}
Honest metric, which was proposed by \cite{nozza2021honest}, is a metric that measures the number of sentence completions that are hurtful in nature. This metric works on both causal models like GPT-2 and masked models like BERT. The prompt for causal models is an incomplete sentence, while for the masked models, it is a sentence with a MASK token which corresponds to the word the will be predicted by the model. The metric computes the percentage of predictions that are unfavorable. The global Honest score is an average of the percentage of unfavorable completions across different semantic classes indicating, semantically negative connotations. and is given by the following equation
\begin{equation}
    \frac{\sum_{s\epsilon class} \sum_{t \epsilon Predictions} isHurtful(t)}{|class|*k}
\end{equation}
In the above equation, s denotes the sentences belonging to a particular class, t denotes the top K predictions of the Language Model. 
\subsection{Log Likelihood}

Log Likelihood metric \cite{nangia-etal-2020-crows} is used to measure how the language model changes its prediction when the context changes. The corpus is initially fed into the model without any modifications and the output of the language model is the prior probability logits. This corpus is later augmented with counterfactual data, such that the only terms changed are the terms that describe the class. The log likelihood is then the probability of the tokens, conditioned on the augmented words. 
\begin{equation}
    Log Score = \sum_{i=0}^{|C|} log P(w_i \epsilon W | W_{\\w_i}, A, \theta)
\end{equation}
In the above equation, w represents the word or the token in the sentence. A is the set of augmented tokens. 
\section{Bias Mitigation Techniques}
In the following sub-sections, we present the bias mitigation methods that are currently supported by Fairpy. 
\subsection{DiffPruning}
Diffpruning is an adversarial training mechanism for mitigating biases in text corpora \cite{hauzenberger2022parameter}. It consists of a set of three neural networks, one for predicting the feature vector, one that predicts the domain or the origin dataset of the feature vector and the the third one which predicts the class label for the instance. 
Each model weight $\theta$ is parameterized as the sum of pretrained weights $\theta_t$ and the mask $\delta$. Finetuning is done by optimizing $\delta$ with L0 Regularization. The $\delta$ term is sparsified with the mask z. The entire process is given by the equation 
\begin{equation}
    \begin{array}{l}
     min_{w_\tau, \alpha_\tau, \beta_\tau} \frac{1}{N}\sum_{n=1}^{N}\mathcal{L}(y_n, m(x_n;\theta + z_\tau \circ w_\tau)) \\
     + \lambda \sum_{i=1}^{d} \sigma (log \alpha_{\tau, i} - \beta_{\tau, i} log (-\gamma / \zeta)
    \end{array}
\end{equation}
\subsection{Null Space Projection}
Iterative Nullspace projection is a technique used to debias the embedding. Given the embedding matrix and a set of protected attributes, the method aims to remove linear dependency between the two using a linear guarding function \cite{liang2021towards}. 
The model is trained to predict the protected attributes from the embedding and this is projected onto the null space of the word embedding to cancel the influence of the embedding on the protected attributes. 
This toolkit uses an autoregressive version of Iterative Nullspace Projection, as proposed in \cite{liang2021towards}. This method performs Null space Projection at every time step t to debias the contextual embedding with respect to a protected attribute. 
\begin{equation}
    \hat{p_\theta}(w_t|c_{t-1}) = \frac{exp(e(w_t)^T*P*f(c_{t-1}))}{\sum_{w\epsilon V}exp(e(w)^T*P*f(c_{t-1}))}
\end{equation}
In the above equation, P is the nullspace of the trained classifier. $f(c_{t-1})$ is the contextual embedding at the previous time step t-1. 
Furthermore, the metric uses a parameter $\alpha$ to determine how to balance the logits of the debiased Language Model with the original one. 
\begin{equation}
    p_\theta(w_t|c_{t-1}) = \alpha \hat{p_\theta}(w_t|c_{t-1}) + (1-\alpha)p*(w_t|c_t-1)
\end{equation} 
where, $p*(w_t|c_{t-1})$ is the predicted of the original Language Model. 

\subsection{Counter Factual Data Augmentation}
Counter Factual Data Augmentation or CDA, is a method that aims to change what the Language Model learns by retraining it on a debiased dataset. Rather than changing the architecture of the model or the embedding structure, this method augments the corpus by replacing all instances of words of a semantic class with another class. This augmented corpus is used for retraining the model. For this metric, the size and the comprehensive nature of the dataset play a big role in the effectiveness of debiasing. The toolkit currently supports the following datasets - Yelp, Reddit and Wikipedia English.

While augmenting the corpus works best for binary settings, it can be extended into multi-class setting by probabilistically replacing the words of a particular class with a member of any other class. While the toolkit currently has a way to support this, this method has not been tested. 
\subsection{Self Debias}
This technique, put forth by \cite{schick2021self} is a way of preventing the generation of biased text by leveraging the internal knowledge of the language model. The debiasing input contains templates of the following type: 

$D_{in}(x,y)$ $\rightarrow$ \textit{The following text 
discriminates} \newline 
\textit{against people because of y; sentence x}

The distributions $P(w | x)$ and $P(w|D_{in}(x,y))$ are calculated. The model is encouraged to produce biased output for the second input. The probability of prediction of unbiased words is left intact but the probability of prediction of biased words is reduced by choosing an appropriate value of $\alpha$ in the below equation
\begin{equation}
    p_M(w|x) = \alpha (\Delta (w, x, y)).p_M(w|x)
\end{equation}
where $\Delta(w, x, y)$ is the difference in the probability distributions of the generated text for the regular input and the debiasing template. It is calculated as follows 
\begin{equation}
    \Delta (w,x,y) = p_M(w | x) - p_M(w | d_{in}(x,y))
\end{equation}
\subsection{Dropout Regularization}
Dropout regularization is a retraining mechanism where the dropout parameters of the language models are tweaked and the models are retrained. This method was first defined by \cite{webster2020measuring}. 
\section{Empirical Analysis}
In this section, we provide an empirical analysis of the bias detection and bias mitigation techniques. The results are presented in Table 1, Table 2 and Table 3. 
The primary findings of the empirical study were as follows. Most metrics were defined in such a way that the prediction probabilities were defined for a word as opposed to tokens. Some of these metrics failed in case of models whose embeddings split the words into smaller tokens, an example was OpenAI-GPT, which failed to generate the StereoSet score because of the internal representation of the model. 

Another case of inconsistency was the final embedding layer of the model. While most models have a final layer that corresponds to the embedding of the output, the way that they are structured and named vary across models, making it hard to provide a single unified interface to perform operations like NullSpace Projection on the models. 

Previous works \cite{meade2021empirical}, have not split the dataset into semantic categories to individually run the metrics against a specific class. We have made attempts to resolve this issue by splitting StereoSet metric and allowing users to run the metric against a single category. This increases the speed and reduces the runtime. Furthermore, the dataset used for measuring WEAT scores - CrowS, has also been split into groups for faster computations, without compromising on the efficiency or accuracy. 

In tables 2 and 3, we present the performance of GPT-2 and BERT after they are debiased with Dropout Regularization based retraining. It can be noted that the WEAT score is the average effect sizes. Both the models show a reduction in the effect size. This is contradictory to the results presented in \cite{meade2021empirical}.

For both the models, counterfactually augmented Yelp-small dataset was used to retrain. 

\begin{table}[h]
\caption{Comparison of bias scores of GPT-2 before and after being debiased with dropout regularization based retraining on augmented Yelp dataset}
\begin{tabular}{lll}
\hline
\textbf{GPT-2 Debiased} & \textbf{Biased} & \textbf{DeBiased} \\
\hline
Hellinger Distance      & 0.14   & 0.35     \\
WEAT Score              & 1.15   & 0.8      \\
StereoSet               & 47.91  & 58.4     \\
Log Probability         & 56.87  & 63.81   \\
\hline
\end{tabular}
\end{table}

\begin{table}[h]
\caption{Comparison of bias scores of BERT before and after being debiased with dropout regularization based retraining on augmented Yelp dataset}
\begin{tabular}{lll}
\hline
\textbf{BERT Debiased}          & \textbf{Biased} & \textbf{Debiased} \\
\hline
Log Probability        & 57.25  & 54.58    \\
F1 Score               & 64.4   & 65.6     \\
StereoSet              & 56.3   & 59.71    \\
WEAT Score             & 1.13   & 0.68    \\
\hline
\end{tabular}
\end{table}
\section{Future Work}
The pipeline forms the basis for debiasing and testing language models for bias and the toolkit currently supports a variety of open source metrics and datasets. However, numerous methods are not included in this toolkit due to compatibility issues and lack of availability. 
The toolkit currently does not support cascading metrics and including hybrid methods that are applied concurrently to the models. This could help us leverage metrics that work in specific setting. 
Lastly, the toolkit would greatly benefit from a web interface that would allow users to plug in their models and extract the statistics from a website. 
\bibliography{acl2020}
\bibliographystyle{acl_natbib}

\end{document}